\documentclass[10pt]{article}

\pdfoutput=1

\usepackage{graphicx}
\graphicspath{{./images/imgplos_pdf/}{./images/}{.}}
\usepackage[pagebackref=true,breaklinks=true,letterpaper=true,colorlinks,bookmarks=false]{hyperref}
\usepackage{rotating}
\usepackage{longtable}
\usepackage{multirow}
\usepackage{multicol}
\usepackage{booktabs}
\usepackage{colortbl}
\usepackage[table]{xcolor}
\usepackage{hhline}
\usepackage{url}
\usepackage[cmex10]{amsmath}
\usepackage{amsbsy}
\usepackage{amssymb}
\usepackage{bm}
\usepackage{algorithm,algorithmic}
\usepackage{scrextend}

\usepackage{authblk}

\date{}

\title{{F-formation Detection: Individuating Free-standing  Conversational Groups in Images}}

\author[1]{Francesco~Setti\thanks{francesco.setti@loa.istc.cnr.it}}
\author[2]{Chris~Russell}
\author[1]{Chiara~Bassetti}
\author[3,4]{Marco~Cristani}
\affil[1]{Institute of Cognitive Science and Technologies (ISTC), Italian National Research Council (CNR), Trento, Italy}
\affil[2]{Department of Computer Science, University College London, UK}
\affil[3]{Department of Computer Science, University of Verona, Italy}
\affil[4]{Pattern Analysis and Computer Vision (PAVIS), Istituto Italiano di Tecnologia Genova(IIT), Italy}

\begin{document}


    \maketitle
	\vspace{-2em}
    \begin{abstract}
      Detection of groups of interacting people is a very interesting and useful task in many modern technologies, with application fields spanning from video-surveillance to social robotics. In this paper we first furnish a rigorous definition of group considering the background of the social sciences: this allows us to specify many kinds of group, so far neglected in the Computer Vision literature. On top of this taxonomy, we present a detailed state of the art on the group detection algorithms. Then, as a main contribution, we present a brand new method for the automatic detection of groups in still images, which is based on a graph-cuts framework for clustering individuals; in particular we are able to codify in a computational sense the sociological definition of F-formation, that is very useful to encode a group having only proxemic information: position and orientation of people. We call the proposed method \emph{Graph-Cuts for F-formation} (GCFF). We show how GCFF definitely outperforms all the state of the art methods in terms of different accuracy measures (some of them are brand new), demonstrating also a strong robustness to noise and versatility in recognizing groups of various cardinality.
    \end{abstract}
	\vspace{2em}


\section{Introduction}
After years of research on automated analysis of individuals, the computer vision community has transferred its attention on the new issue of modeling gatherings of people, commonly referred as \emph{groups}~\cite{gatica2009automatic,Cristani:SISM:CVPRW:2010,Cristani:FF:BMVC:2011,ge2012vision}.

A group can be broadly understood as a social unit comprising several members who stand in status and relationships with one another~\cite{forsyth2010group}. However, there are many kinds of groups, that differ in dimension (small groups or crowds), durability (ephemeral, \emph{ad hoc} or stable groups), in/formality of organization, degree of ``sense of belonging'', level of physical dispersion etc.~\cite{goffman1961encounters} (see the literature review in the next section).
In this article, we build from the concepts of sociological analysis and we focus on free-standing conversational groups (FCGs), or small ensembles of co-present persons engaged in \emph{ad hoc} focused encounters~\cite{goffman1961encounters,goffman1966behavior,kendon1990conducting}.
FCGs represent crucial social situations, and one of the most fundamental bases of dynamic sociality: these facts make them a crucial target for the modern automated monitoring and profiling strategies which have started to appear in the literature in the last three years~\cite{Cristani:FF:BMVC:2011,Cristani:SocialExpert:ExpSys:2012,hung2011detecting,setti2013group,Cristani:FF:ICIP:2013,tran2013social}.

In computer vision, the analysis of groups has occurred historically in two broad contexts: video-surveillance and meeting analysis.

Within the scope of video-surveillance, the definition of a group is generally  simplified to two or more people of similar velocity, spatially and temporally close to one another~\cite{garate:hal-00879734}. This simplified definition arises from the difficulty of inferring persistent social structure from short video clips.
In this case, most of the vision-based approaches perform group tracking, i.e. capturing individuals in movement and maintaining their identity across video frames, understanding how they are partitioned in groups~\cite{lau2010multi,ge2012vision,Qin:CVPR2012,Bazzani:CVPR12,mazzon2013detection,garate:hal-00879734}. 

In meeting analysis, typified by classroom behavior~\cite{gatica2009automatic}, people typically sit around a table and remain near a fixed location for most of the time, predominantly interacting through speech and gesture. In such a scenario, activities can be finely monitored using a variety of audiovisual features, captured by pervasive sensors like portable devices, microphone arrays, etc.~\cite{zhang2004modeling,jayagopi2009modeling,hung2010estimating}.

From a sociological point of view, meetings are examples of social organization that employs \emph{focused} interaction, which occurs when persons openly cooperate to sustain a single focus of attention~\cite{goffman1961encounters,goffman1966behavior}. This broad definition covers other collaborative situated systems of activity that entail a more or less static spatial and proxemic organization -- such as playing a board or sport game, having dinner, doing a puzzle together, pitching a tent, or free conversation~\cite{goffman1961encounters}, whether  sitting on the couch at a friend's place, standing in the foyer and discussing the movie, or leaning on the balcony and smoking a cigarette during work-break.

Free-standing conversational groups (FCGs)~\cite{kendon1990conducting} are another example of \emph{focused} encounters. FCGs emerge during many and diverse social occasions, such as a party, a social dinner, a coffee break, a visit in a museum, a day at the seaside, a walk in the city plaza or at the mall; more generally, when people spontaneously decide to be in each other's immediate presence to interact with one another.
For these reasons, FCGs are fundamental social entities, whose automated analysis may bring to a novel level of activity and behavior analysis.

In a  FCG, people communicate to the other participants, among --and above all-- the rest, what they think they are doing together, what they regard as the activity at hand.
And they do so not only, and perhaps not so much, by talking, but also, and as much, by exploiting non-verbal modalities of expression, also called social signals \cite{vin09}, among which positional and orientational forms play a crucial role (cf. also~\cite{goffman1966behavior}, p. 11).
In fact, the spatial position and orientation of people define one of the most important proxemic notions which describe an FCG, that is, the Adam Kendon's \emph{Facing Formation}, mostly known as \emph{F-formation}.

\begin{figure}[!h]
  \centering
  \includegraphics[width=0.5\columnwidth]{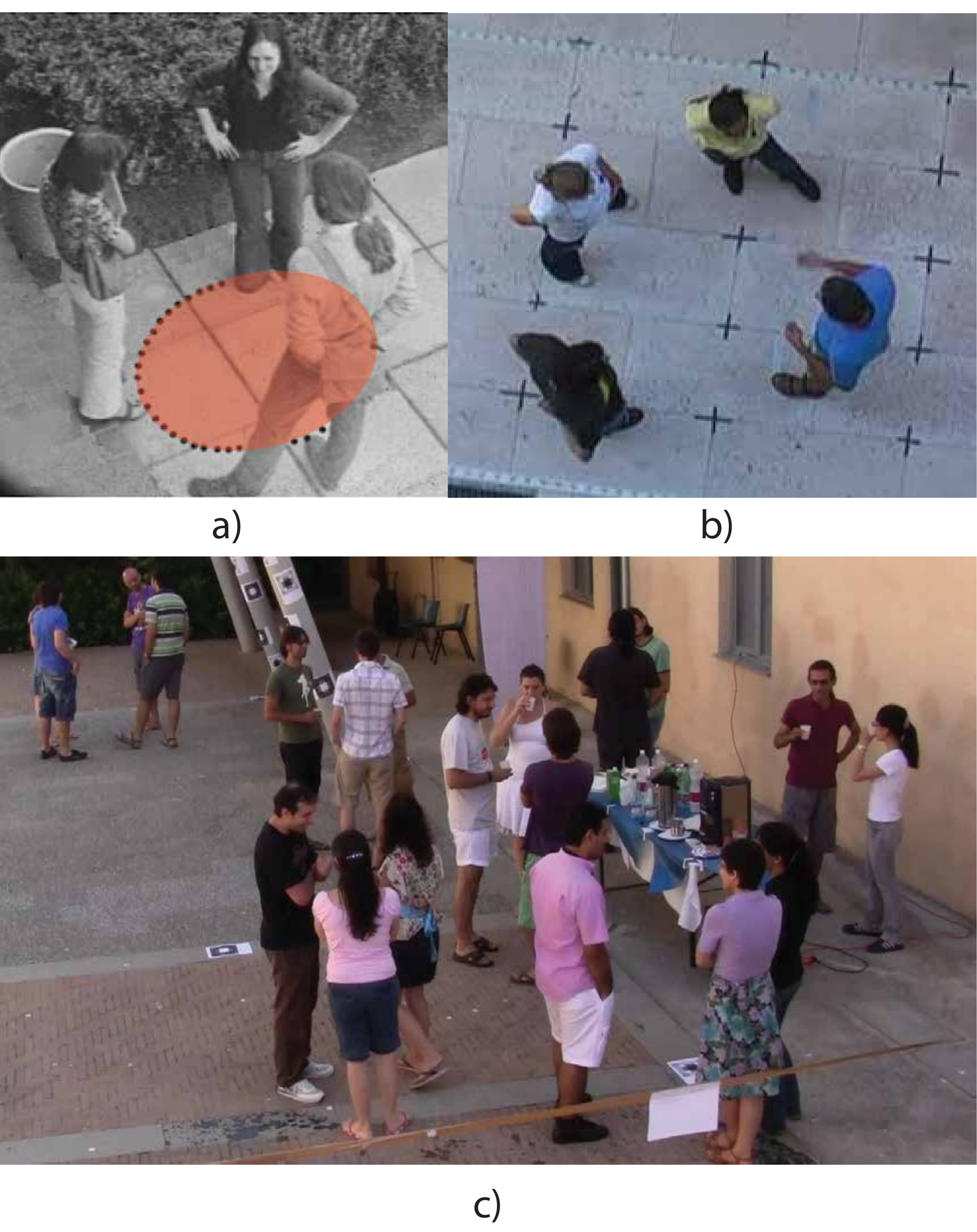}
  \caption{{\bf Examples of F-formations.} a) in orange, the o-space; b) an aerial image of a circular F-formation; c) a party, something similar to a typical surveillance setting with the camera located 2-3 meters from the floor: detecting F-formations here is challenging.}
  \label{fig:examples}
\end{figure}

In Kendon's terms~\cite{kendon199213,ciolek1980environment,kendon1990conducting}, an F-formation is a socio-spatial formation in which people have established and maintain a convex space (called \emph{o-space}) to which everybody in the gathering has direct, easy and equal access. Typically, people arrange themselves in a form of a circle, ellipse, horseshoe, side-by-side or L-shape (cf. Fig.~\ref{fig:F-formations}), so that they can have easy and preferential access to one another while excluding distractions of the outside world with their backs. Examples of F- formations are reported in Fig.~\ref{fig:examples}. In computer vision, spatial position and orientational information can be automatically extracted, and these facts pave the way to the computational modeling of F-formation and, as a consequence, of the FCGs.

Detecting free-standing conversational groups is useful in many contexts. In video-surveillance, automatically understanding the network of social relationships observed in an ecological 
scenario may result beneficial for advanced suspect profiling, improving and automatizing SPOT (Screening Passengers by Observation Technique) protocols~\cite{florence2009profiles}, which nowadays are performed uniquely by human operators.

A robust FCG detector may also impact the social robotics field, where the approaches so far implemented work on few number of people, usually focusing on a single F-formation~\cite{huttenrauch2006investigating,yousuf2012development,nieuwenhuisen2013human}.

Efficient identification of  FCGs could be of use in multimedia applications, and especially in semantic tagging~\cite{gallagher2009understanding,Marin11}, where groups of people are currently inferred by the proximity of their faces in the image plane. Adopting systems for 3D pose estimation from 2D images~\cite{andriluka2012human} plus an FCG detector could in principle lead to more robust estimations. In this scenario, the extraction of social relationships could help in inferring personality traits~\cite{raacke2008myspace,gosling2007personality} and triggering friendship invitation mechanisms~\cite{kosinski2013private}.

In computer-supported cooperative work (CSCW), being capable of automatically detecting FCG could be a step ahead in understanding how computer systems can support socialization and collaborative activities: e.g., ~\cite{suzuki1995interaction,morrison2008electronic,marshall2011using,Akpan:2013:EES:2470654.2481306}; in this case, FCGs are usually found by hand, or employing wearable sensors.

Manual detection of FCGs occurs also in human computer interaction, for the design of devices reacting to a situational change~\cite{hornecker2005design,schnadelbach2012hybrid}: here the benefit of the automation of the detection process may lead to a genuine systematic study of how proxemic factors shape the usability of the device.

The last three years have seen works  that automatically detect F-formations: Bazzani et al.~\cite{Cristani:SocialExpert:ExpSys:2012} first proposed the use of positional and orientational information to capture Steady Conversational Groups (SCG); 
Cristani et al.~\cite{Cristani:FF:BMVC:2011} designed a sampling technique to seek F-formations centres by performing a greedy maximization in a Hough voting space; Hung and Kr{\"o}se~\cite{hung2011detecting} detected F-formations by finding distinct maximal cliques in weighted graphs via graph-theoretic clustering; both the techniques were compared by Setti et al.~\cite{setti2013group}. A multi-scale extension of the Hough-based approach~\cite{Cristani:FF:BMVC:2011} was proposed by Setti et al.~\cite{Cristani:FF:ICIP:2013}. This improved on previous works, by explicitly modeling F-formations of different cardinalities. Tran et al.~\cite{tran2013social} followed the graph based approach of~\cite{hung2011detecting}, extending it to deal with video-sequences and recognizing five kinds of activities.\\

Our proposed approach detects an arbitrary number of F-formations on single images using a monocular camera, by considering as input the position of people on the ground floor, and their orientation, captured as the head and/or body pose. The approach is iterative, and starts by assuming an arbitrarily high number of F-formations: after that, a hill-climbing optimisation alternates between assigning individuals to F-formations using the efficient graph-cut based optimisation~\cite{co-oc}, and updating the centres of the F-formations, pruning unsupported groups in accordance with a Minimum Description Length prior. The iterations continue until convergence, which is guaranteed.


As a second contribution, we present a novel set of metrics for group detection. This is not constrained to apply to FCG, but to any set of people considered as a whole, thus embracing generic group or crowd tracking scenarios~\cite{tran2013social}.

%
The fundamental idea is the concept of \emph{tolerance threshold}, which basically regulates the tolerance on individuating groups, allowing some individual components to be missed or external people to be added in a group. Thanks to the tolerance threshold, the concepts of \emph{tolerant match}, \emph{tolerant accuracy} and of \emph{precision} and \emph{recall} can be easily derived. Such measures take inspiration from the group match definition, firstly published in a previous work~\cite{Cristani:FF:BMVC:2011} and adopted in many recent group detection~\cite{tran2013social,Cristani:FF:ICIP:2013} and group tracking methods~\cite{Bazzani:CVPR12} so far: in practice, it corresponds to fix the tolerance threshold to a predefined value.

In this article, we show that, by letting the tolerance threshold change in a continuous way from maximum to minimum tolerance, it is possible to get an informative and compact measure (in the form of area under the curve) that summarises the behaviour of a given detection methodology. In addition, the tolerant match can be applied specifically to groups of a given cardinality, allowing to obtain specific values of accuracy, precision and recall; this highlights the performance of a given approach in a specific scenario, that is, the ability of capturing small or large groups of people.\\

\begin{figure}[!h]
  \centering
  \includegraphics[width=.9\columnwidth]{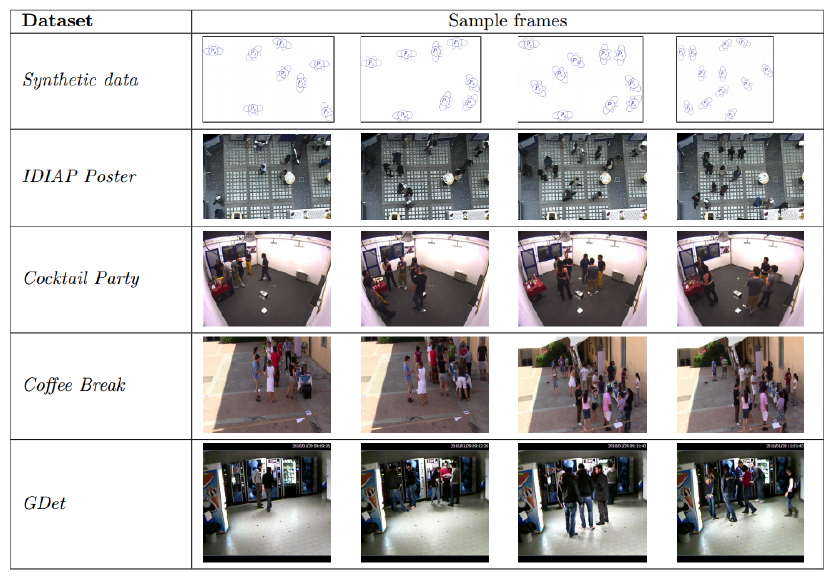}
  \caption{{\bf Sample images of the four real-world datasets.} For each dataset four frames are reported showing different situations of crowd and arrangement.}
  \label{fig:eximages}
\end{figure}

In the experiments, we apply GCFF to all publicly available 
datasets (see Fig.~\ref{fig:eximages}), 
consisting of more than 2000 different F-formations over 1024 frames. Comparing against the five most accurate methods in the literature we definitely set the best score on every dataset. 
In addition, using our novel metrics, we show that GCFF has the best behaviour in terms of robustness to noise, and it is able to capture groups of different cardinalities without changing any threshold.
Summarising, the main contributions of this article are the following:

\begin{itemize}
 \item  A novel methodology to detect F-formations
from single images acquired by a monocular camera, which operates on positional and orientational information of the individuals in the scene. %
Unlike previous approaches, our novel methodology is a direct formulation of the sociological principles (proximity, orientation and ease of access) concerning o-spaces.
The strong conceptual simplicity and clarity of our approach is an asset in two important ways: we do not require bespoke optimisation techniques, and 
we make use of established methods known to work reliably and efficiently. Second, and by far more important, the high accuracy and clarity of our approach, along with its basis in sociological principles makes it well suited for use in the social sciences as means of automatically annotating data.
 \item A rigorous taxonomy of the group entity, which takes from social science and illustrates all the different group manifestations, delineating their main characteristics, in order to go beyond the generic term of group, often misused in the computer vision community.
 \item A novel set of metrics for group detection, that for the first time models the fact that a group could be partially captured, with some people missing or erroneously taken into account, through the concept of tolerant match. %
The metrics can be employed to whatever approach involving groups (group tracking included).
\end{itemize}

The remainder of the paper is organised as follows: the next section presents a literature review of group modeling, with particular emphasis on the terminology adopted, which will be imported from the social and cognitive sciences; the proposed GCFF approach, together with its sociological grounding, is presented afterwards, followed by an extensive experimental evaluation. Finally, we will draw the conclusion and envisage the future perspectives.

---

\section{Literature Review}\label{sec:SoA}
Research on group modeling in computer science is highly multidisciplinary, necessarily encompassing the social and the cognitive sciences when it comes to analyse human interaction. In this multifaceted scenario, characterising the works most related to our approach requires us to distinguish between related sociological concepts; starting with the Goffmanian~\cite{goffman1961encounters} notions, of (a) ``group'' vs. ``gathering'', (b) ``social occasion'' vs. ``social situation'', (c) ``unfocused'' vs. ``focused'' interaction, and (d) Kendon's~\cite{kendon1988goffman} specification concerning ``common focused'' vs. ``jointly focused'' encounters.

 As mentioned in the introduction, \emph{groups} entail some durable membership and organisation, \emph{gatherings} consist of any set of two or more individuals in mutual immediate presence at a given moment. When people are co-present, they tend to behave like one who participates in a social occasion, and the latter provides the structural social context, the general ``scheme'' or ``frame'' of behaviour --like a party, a conference dinner, a picnic, an evening at the theatre, a night in the club, an afternoon at the stadium, a walk together, a day at the office, etc.-- within which gatherings (may) develop, dissolve and redevelop in diverse and always different situational social contexts (or social situations, that is, e.g., that specific party, dinner, picnic, etc.)~\cite{goffman1966behavior}.

\emph{Unfocused interaction} occurs whenever individuals find themselves by circumstance in the immediate presence of others. For instance, when forming a queue or crossing the street at a traffic light junction. On such occasions, simply by virtue of the reciprocal presence, some form of interpersonal communication must take place regardless of individual intent. Conversely, \emph{focused interaction} occurs whenever two or more individuals willingly agree --although such an agreement is rarely verbalised-- to sustain for a time a single focus of cognitive and visual attention~\cite{goffman1961encounters}. Focused gatherings can be further distinguished in \emph{common focused} and \emph{jointly focused} one~\cite{kendon1988goffman}. %
The latter entails the sense of a mutual, instead of merely common, activity; a preferential openness to interpersonal communication, an openness one does not necessarily find among strangers at the theatre, for instance; in other words, a special communication license, like in a conversation, a board game, or a joint task carried on by a group of face-to-face interacting collaborators. Participation, in other words, is not at all peripheral but engaged; people are -- and display to be -- mutually involved~\cite{goffman1966behavior}. All this can exclude from the gathering others who are present in the situation, as in any FCG at a coffee break with respect to the other ones.

\begin{figure}[!h]
  \centering
  \includegraphics[width=.9\columnwidth]{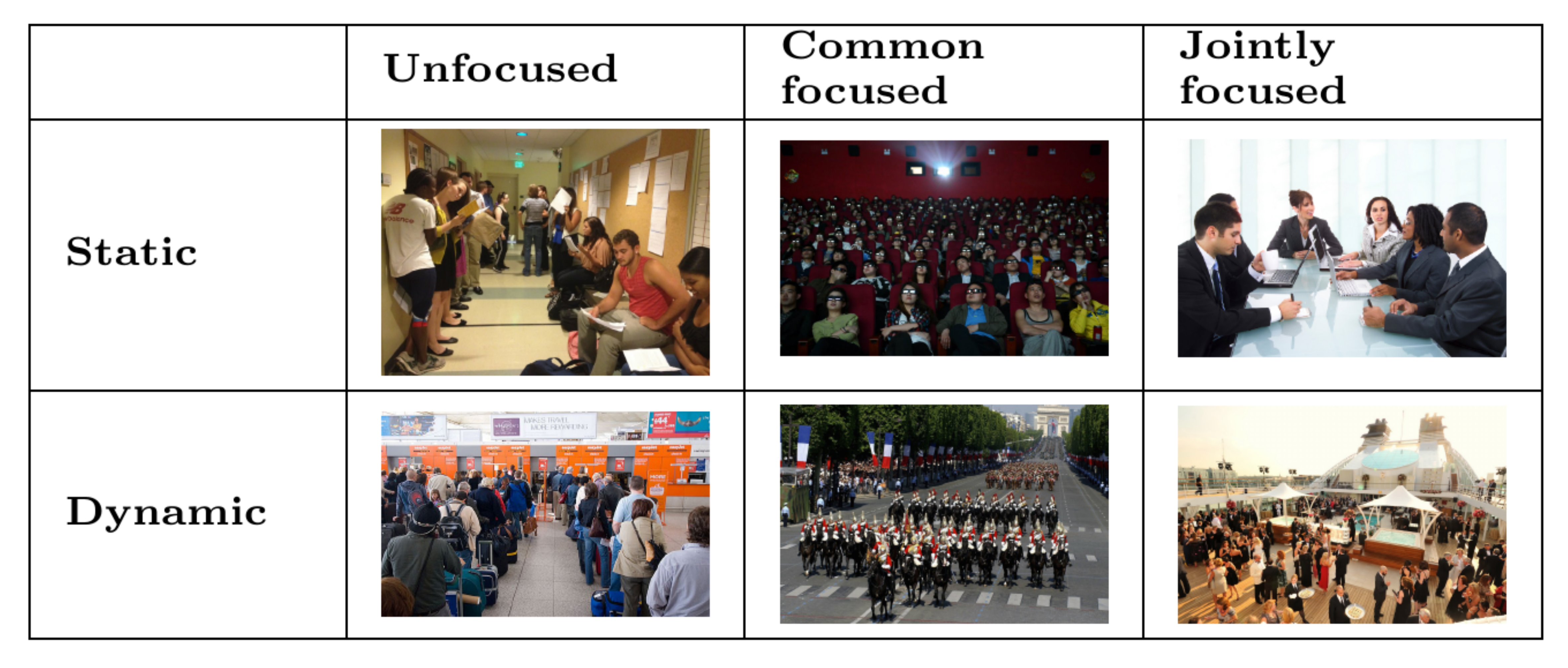}
  \caption{{\bf Examples of gatherings categorised by focus of attention and spatio-proxemic freedom.} Jointly focused, dynamic: our case, FCGs at a cocktail party; common focused, dynamic: a parading platoon; unfocused, dynamic: a queue at the airport; jointly focused, static: a meeting; common focused, static: people in a theatre stand; unfocused, static: persons in a waiting room.}
  \label{fig:SoA_images}
\end{figure}

Finally, we should consider the static/dynamic axis concerning the degree of freedom and flexibility of the spatial, positional, and orientational organisation of gatherings. Sometimes, indeed, people maintain approximately their positions for an extended period of time within fixed physical boundaries (e.g., during a meeting); sometimes they move within a delimited area (e.g., at a party); and sometimes they do within a more or less unconstrained  space (for instance, people conversing while walking in the street). It is about a continuum, in which we can analytically identify thresholds.
Tab.~\ref{tab:examples} lists some categorised examples of gatherings, considering the taxonomy axis ``static/dynamic organisation'' and the ``unfocused/common-focused/jointly-focused interaction'' one.
Fig.~\ref{fig:SoA_images} shows some categorised examples of encounters.

\begin{table}[h!]
  \centering
  \scriptsize
  \begin{tabular}{|p{.13\textwidth}|p{.26\textwidth}|p{.26\textwidth}|p{.26\textwidth}|}
    \hline
     & \textbf{Unfocused} & \textbf{Common focused} & \textbf{Jointly focused} \\
    \hline
    \textbf{Static} & open-space offices & conferences, classrooms & meetings, board-game play\\
    \hline
    \textbf{Dynamic} & waiting rooms, queues, promenades, airports, stations, street-crossings
 & theatre stands, stadium stands, parades, processions, demonstrations
 & restaurants, having meal together, work-breaks, bar, clubs, pubs\\
    \hline
  \end{tabular}
  \caption{Gatherings categorisation on the basis of focus of attention and spatio-proxemic freedom exemplified by typical social settings/situations.}
  \label{tab:examples}
\end{table}

Within this taxonomy, our interest is on \emph{gatherings}, formed by people \emph{jointly focused} on interacting 
in a \emph{quasi-static} fashion within a dynamic scenario. Kendon dubbed this scenario as characterising \emph{free-standing conversational groups}, highlighting their spontaneous aggregation/disgregation nature, implying that their members are jointly focused, and specifying their mainly-static proxemic layout within a dynamic proxemic context.

The following review centres on the case of FCGs and their formation, while for the other cases we refer: with respect to computer vision, to~\cite{aggarwal2011human} for generic human activity analysis, including single individuals, groups and crowds, and to~\cite{junior2010crowd} for a specific survey on crowds; with respect to the sociological literature, to~\cite{goffman1966behavior} as for unfocused gatherings, to~\cite{kendon1988goffman,schweingruber1999method} as for common focused ones, and to~\cite{kaya2007territoriality,mcphai1994clusters} as for crowds in particular.

The analysis of focused gatherings in computer science had the first traces appearing in the field of human computer interaction and robotics, especially for what concerns context- aware computing, computer-supported cooperative work and social robotics~\cite{hornecker2005design,schnadelbach2012hybrid,ballendat2010proxemic,jungmann2014spatial}.
This happened since the detection of focused gatherings requires finer feature analysis, and in particular body posture inference other than positional cues extraction: these are difficult tasks for traditional computer vision scenarios, where people is captured at low resolution, under diverse illumination conditions, often partially or completely occluded.

In human-computer interaction, F-formation analysis encompasses context-aware computing, by considering spatial relationships among people where space factors become crucial into the design of applications for devices reacting to a situational change~\cite{hornecker2005design,schnadelbach2012hybrid}. In particular, Ballendat et al.~\cite{ballendat2010proxemic} studied how proxemic interaction is expressive when considering cues like position, identity, movement, and orientation. They found that these cues can mediate the simultaneous interaction of multiple people as an F-formation, interpreting and exploiting people's directed attention to other people. So far, the challenge with these applications for researchers has been the hardware design, while the social dynamics are typically not explored. As notable exception, Jungman et al.~\cite{jungmann2014spatial} studied how different kinds of F-formations (L-shaped vs. face-to-face) identify different kinds of interaction: in particular, they examined whether or not Kendon's observation according to which face-to-face configurations are preferred for competitive interactions whereas L-shaped configurations are associated with cooperative interactions holds in gaming situations. The results partially supported the thesis.

In computer-supported cooperative work, Suzuki and Kato~\cite{suzuki1995interaction} described how different phases of collaborative working were locally and tacitly initiated, accomplished and closed by children by moving back and forth between standing face-to-face formations and sitting screen-facing formations. Morrison et al.~\cite{morrison2008electronic} studied the impact of the adoption of electronic patient records on the structure of F-formations during hospital ward rounds. Marshall et al.~\cite{marshall2011using} analysed through F-formations the social interactions between visitors and staff in a tourist information centre, describing how the physical structures in the space encouraged and discouraged particular kinds of interactions, and discussing how F-formations might be used to think about augmenting physical spaces. Finally, Akpan et al.~\cite{Akpan:2013:EES:2470654.2481306}, for the first time, explored the influence of both physical space and social context (or place) on the way people engage through F-formations with a public interactive display. The main finding is that social properties are more important than merely spatial ones: a conducive social context could overcome a poor physical space and encourage grouping for interaction; conversely, an inappropriate social context could inhibit interaction in spaces that might normally facilitate mutual involvement.
So far, no automatic F-formation detection has been applied: positional and orientational information were analysed by hand, while our method is fully automated.

In social robotics, Nieuwenhuisen and Behneke presented Robotinho~\cite{nieuwenhuisen2013human}, a robotic tour guide which resembles behaviour of human tour guides and leads people to exhibits in a museum, asking people to come closer and arrange themselves in an F-formation, such that it can attend the visitors adequately while explaining an exhibit. Robotinho detects people by first detecting their faces, and using laser-range measurements to detects legs and trunks. Given this, it is not clear how proper F-formations are recognised. Robotinho essentially improves what has been done by Yousuf et al.~\cite{yousuf2012development}, that develop a robot that simply detect when an F-formation is satisfied before explaining an exhibit.
In this case, F-formations were detected automatically, using advanced sensors (range cameras, etc.) with the possibility of checking just one formation. In our case, a single monocular camera is adopted and the number of F-formations is not bounded.

In computer vision, Groh et al.~\cite{groh2010detecting_3} proposed to use the relative shoulder orientations and distances (using wearable sensors) between each pair of people as a feature vector for training a binary classifier, learning the pairwise configurations of people in a FCG and not. Strangely, the authors discouraged large FCG during the data acquisition, introducing a bias on their cardinality. With our proposal, no markers or positional devices have been considered, and entire FCGs of arbitrary cardinality are found (not pairwise associations only).
In his previous work~\cite{Cristani:SocialExpert:ExpSys:2012}, one of the authors started to analyse F-formations by checking the intersection of the view-frustum of neighbouring people, where the view frustum was automatically detected by inferring the head orientation of each single individual in the scene. Under a sociological perspective, the head orientation cue can be exploited as an approximation of a person's focus of visual and cognitive attention, which in turn acts as an indication of the body orientation and the foot position, the last one considered as the most proper way to detect F-formations. Hung and Kr\"{o}se~\cite{hung2011detecting} proposed to consider an F-formation as a dominant-set cluster~\cite{pavan2007dominant} of an edge-weighted graph, where each node in the graph is a person, and the edges between them measure the affinity between pairs. Such maximal cliques has been defined by Pavan and Pelillo as dominant sets~\cite{pavan2007dominant}, for which a game theoretic approach has been designed to solve the clustering problem under these constraints. More recently, Tran et al.~\cite{tran2013social} applied a similar graph-based approach for finding groups, which were subsequently analysed by a specific descriptor that encodes people's mutual poses and their movements within the group  gathering for activity recognition.
In all these three approaches, the common underlying idea is to find set of pairs of individuals with similar mutual pose and orientation, thus considering pairwise proxemics relations as basic elements. This is weak, since in practice it tends to find circular formations (that is, cliques with compact structures), while FCGs have other common layouts (side-by- side, L-shape, etc.). In our case, all kinds of F-formations can be found. In addition, the definition of F-formation requires that no obstacles must invade the o-space (the convex space surrounded by the group members, see Fig.~\ref{fig:examples}a): whereas in the above-mentioned approaches such a condition is not explicitly taken into account, it is a key element in GCFF.

In this sense, GCFF shares more similarities with the work of Cristani et al.~\cite{Cristani:FF:BMVC:2011}, where F-formations were found by considering as atomic entity the state of a single person: each individual projects a set of samples in the floor space, that vote for different o-space centres, depending on his or her  position and orientation. Votes are then accumulated in a proper Hough space, where a greedy minimization finds the subset of people voting for the same o-space centre, which in turns is free of obstacles. Setti et al.~\cite{setti2013group} compared the Hough-based approach with the graph-based strategy of Hung and Kr\"{o}se~\cite{hung2011detecting}, finding that the former performs better, especially when in presence of high noise. The study was also aimed at analysing how important positional and orientational information are: it turned out that, when in presence of positional information only, the performances of the Hough-based approach decrease strongly, while graph-based approaches are more robust. Another voting-based approach resembling the Hough-based strategy has been designed by Gan et al.~\cite{Gan:2013:TEF:2502081.2502096}, who individuated a global interaction space as the overlap area of different individual interaction spaces, that is, conic areas aligned coherently with the body orientations of the interactants (detected using a kinect device). Subsequently, the Hough-based approach has been extended for dealing with groups of diverse cardinalities by Setti et al.~\cite{Cristani:FF:ICIP:2013}, who adopted a multi-scale Hough-space, and set the best performance so far.

\section{Method}\label{sec:Our}
\begin{figure*}[!h]
  \centering
  \includegraphics[width=.9\textwidth]{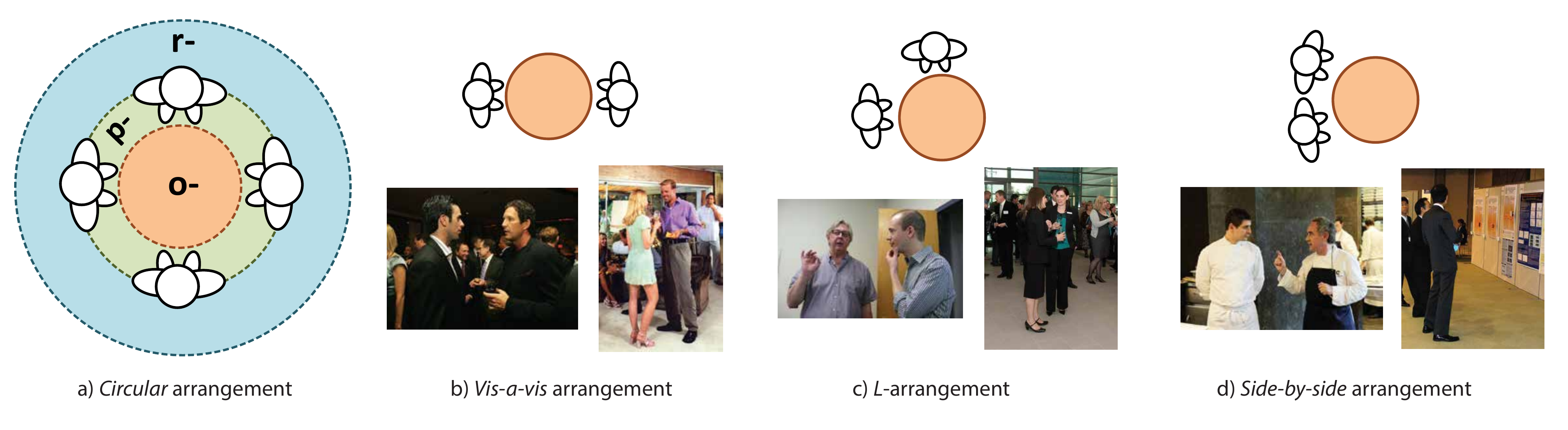}
  \caption{{\bf Structure of an F-formation and examples of F-formation arrangements.} a) Schematization of the three spaces of an F-formation: starting from the centre, \emph{o-space}, \emph{p-space} and \emph{r-space}. b-d) Three examples of F-formation arrangements: for each one of them, one picture highlights the head and shoulder pose, the other shows the lower body posture. For a picture of circular F-formation, see also Fig.~\ref{fig:examples}.}
  \label{fig:F-formations}
\end{figure*}

Our approach is strongly based on the formal definition of F-formation given by Kendon~\cite{kendon1990conducting} (\emph{page 209}):
\begin{quote}
\emph{An F-formation arises whenever two or more people sustain a spatial and orientational relationship in which the space between them is one to which they have equal, direct, and exclusive access.}
\end{quote}
In particular, an F-formation is the proper organisation of three social spaces: \emph{o-space}, \emph{p-space} and \emph{r-space} (see Fig.~\ref{fig:F-formations}a).

The o-space is a convex empty space surrounded by the people involved in a social interaction, where every participant is oriented inward into it, and no external people are allowed to lie. 
More in the detail, the o-space is determined by the overlap of those regions dubbed \emph{transactional segments}, where as transactional segment we refer to the area in front of the body 
that can be reached easily, and where hearing and sight are most effective~\cite{ciolek1983proxemics}. In practice, in a F-formation, the transactional segment of a person coincides with the o-space, and this fact has been exploited in our algorithm.
The p-space is the belt of space enveloping the o-space, where only the bodies of the F-formation participants (as well as some of their belongings) are placed. People in the p-space participate to an F-formation using the o-space to transmit their messages.
The r-space is the space enveloping o- and p-spaces, and is also monitored by the F-formation participants. People joining or leaving a given F-formation mark their arrival as well as their departure by engaging in special behaviours displayed in a special order in special portions of r-space, depending on several factors (context, culture, personality among the others); therefore, here we prefer to avoid the analysis of such complex dynamics, leaving their computational analysis as future work.\\

F-formations can be organised in different \emph{arrangements}, that is, spatial and orientational layouts (see Fig.~\ref{fig:F-formations}a-d)~\cite{cook1970experiments,ciolek1980environment,kendon1990conducting}. In F-formations of two individuals, usually we have a \emph{vis-a-vis} arrangement, in which the two participants stand and face one another directly; another situation is the \emph{L-arrangement}, when two people lie in a right angle to each other. As studied by  Kendon~\cite{kendon1990conducting}, vis-a-vis configurations are preferred for competitive interactions, whereas L-shaped configurations are associated with cooperative interactions. In a \emph{side-by-side} arrangement, people stand close together, both facing the same way; this situation occurs frequently when people stand at the edges of a setting against walls.
\emph{Circular} arrangements, finally, hold when F-formations are composed by more than two people; other than being circular, they can assume an approximately linear,
semicircular, or rectangular shape.

GCFF finds the o-space of an F-formation, assigning to it those individuals whose transactional segments do overlap, without focusing on a particular arrangement.
Given the position of an individual, to identify the transactional segment we exploit orientational information, which may come from the head orientation, the shoulder orientation or the feet layout, in increasing order of reliability~\cite{kendon1990conducting}. The idea is that the feet layout of a subject indicates the mean direction along which his messages should be delivered
, while he is still free to rotate his head and to some extent his shoulders through a considerable arc, before he must begin to turn his lower body as well. The problem is that feet are almost impossible to detect in an automatic fashion, due to the frequent (auto) occlusions; shoulder orientation is also complicated, since most of the approaches of body pose estimation work on 2D data and do not manage auto-occlusions. However, since any sustained head orientation in a given direction is usually associated with a reorientation of the lower body (so that the direction of the transactional segment again coincides with the direction in which the face is oriented~\cite{kendon1990conducting}), head orientation should be considered proper for detecting transactional segments and, as a consequence, the o-space of an F-formation. In this work, we assume to have as input both positional information and head orientation; this assumption is reasonable due to the massive presence of robust tracking technologies~\cite{benfold2011stable} and head orientation algorithms~\cite{10.1109/TPAMI.2007.70773,ba2011multiperson,chen2012we}.

In addition to this, we consider soft exclusion constraints: in an o-space, F-formation participants should have \emph{equal, direct and exclusive access}. In other words, if person $i$ stands between another person $j$, and an o-space centre $O_g$ of the F-formation $g$, this should prevent $j$ from focusing on the o-space, and, as a consequence, from being part of the related F-formation.

\begin{figure}[!h]
  \centering
  \includegraphics[width=0.4\textwidth]{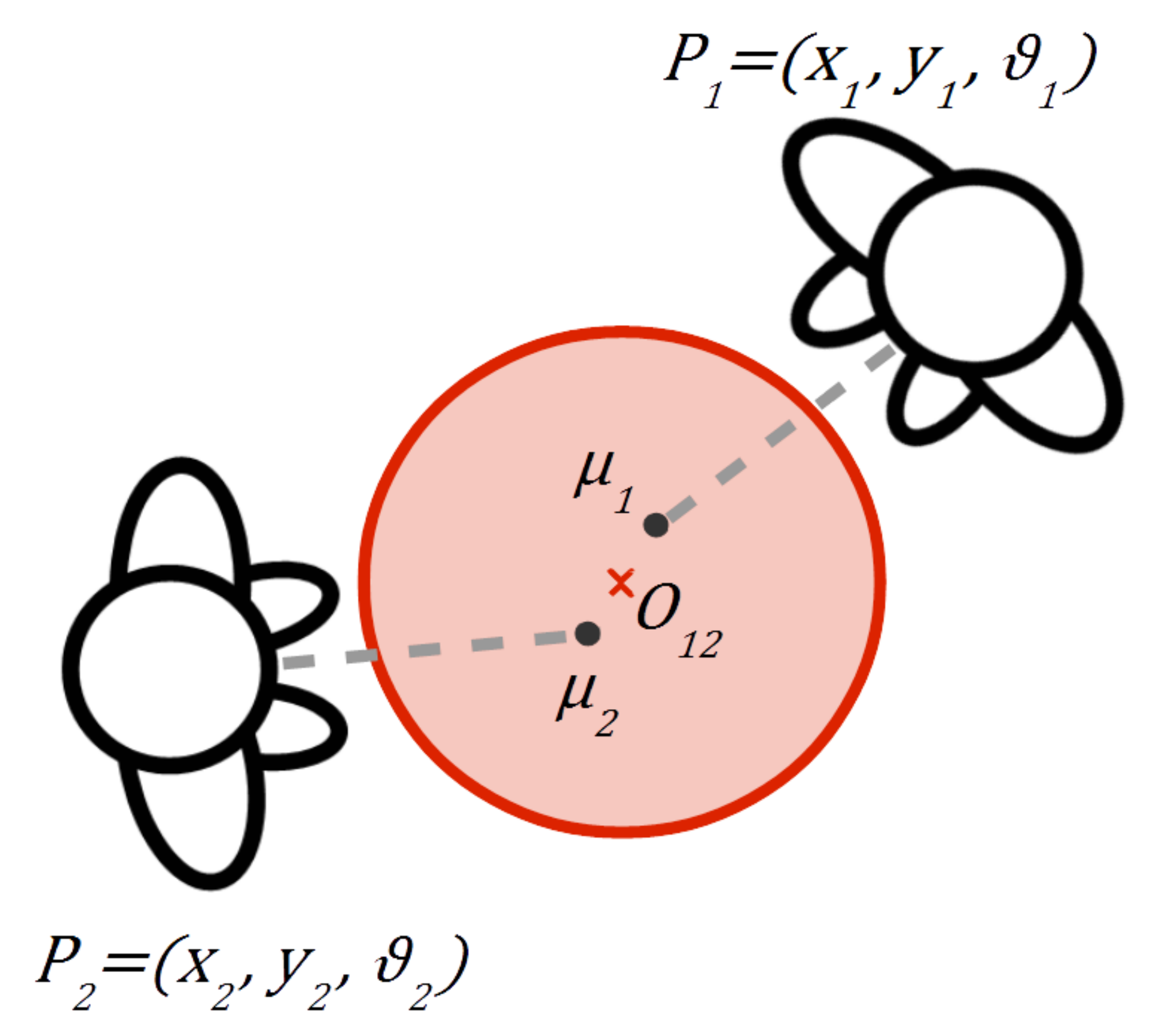}
  \caption{{\bf Schematic representation of the problem formulation.} Two individuals facing each other, the gray dot representing the transitional segment centre, the red cross being the o-space centre and the red area the o-space of the F-formation.}
  \label{fig:fformschema}
\end{figure}

In what follows, we formally define the objective function accounting for positional, orientational and exclusion constraints aspects, and show how it can be optimised.
Fig.~\ref{fig:fformschema} gives a graphical idea of the problem formulation.

\subsection{Objective Function}\label{sec:graphbuild}
We use $P_i = [x_{i},y_{i},\theta_i]$ to represent the position
$x_{i},y_{i}$ and head orientation $\theta_i$ of the individual $i
\in \{1,\ldots,n\}$ in the scene. Let $TS_i$ be the a priori distribution which models the transactional segment of individual $i$. As we explained in the previous section, this segment is coherent with the position and orientation of the head, so we can assume $TS_i \sim \mathcal{N}(\mu_i,\mathbf{\Sigma}_i)$, where $\mu_i=[x_{\mu_i},y_{\mu_i}]=[x_{i}+ D \cos \theta_i ,y_{i}+ D \sin \theta_i]$,  $\mathbf{\Sigma}_i=\sigma\cdot\mathbf{I}$ with $\mathbf{I}$ the 2D identity matrix, and $D$ is the distance between the individual $i$ and the centre of its transactional segment (hereafter called \emph{stride}).
The stride parameter $D$ can be learned by cross-validation, or fixed a priori accounting for social facts. In practice, we assume the transactional segment of a person having a circular shape, which can be thought as superimposed to the o-space of the F-formation she may be part of.

$O_g = [u_{g},v_{g}]$ indicates the position of a candidate o-space centre for F-formation $g \in \{1,M\} $, while we use $G_i$ to refer to the F-formation
containing individual $i$, considering the F-formation assignment $G_i=g$ for some $g$. The assignment assumes that each individual $i$ may belong to a single F-formation $g$ only\footnote{For the sake of mathematical simplicity, we assume that each lone individual not belonging to a gathering can be considered as a \emph{spurious} F-formation.} at any given time, and this is reasonable when we are focusing one a single time, that is, an image.
It follows naturally the definition of $O_{G_i}=[u_{G_i},v_{G_i}]$, which represents the position of a candidate o-space centre for an unknown F-formation $G_i=g$ containing $i$.

At this point, we define the likelihood
probability of an individual $i$'s transitional segment centre $C_i=[u_i ,v_i ]$ given the a priori variable $TS_i$.
\begin{align}
\Pr(C_i|TS_i)&\propto\exp\left(-\frac{||C_i - \mu_i||_2^2}{\sigma^2}\right)\\
&=\exp\left(-\frac{(u_i - x_{\mu_i})^2+(v_i-y_{\mu_i})^2}{\sigma^2}\right)
\end{align}

Hence, the probability that an individual $i$ shares an o-space centre $O_{G_i}$ is given by
\begin{equation}
\Pr(C_i=O_{G_i}|TS_i)
\propto\exp\left(-\frac{(u_{G_i}-x_{\mu_i})^2+(v_{G_i}-y_{\mu_i})^2}{\sigma^2}\right)
\end{equation}
and the posterior probability of any overall assignment is given by
\begin{equation}
\Pr(C=O_{G}|TS)\propto\prod_{i \in [1,n]}\exp\left(-\frac{(u_{G_i}-x_{\mu_i})^2+(v_{G_i}-y_{\mu_i})^2}{\sigma^2}\right)
\end{equation}
with $C$ the random variable which models a possible joint location of all the o-space centres, $O_{G}$ is one instance of this joint location, and $TS$ is the position of all the transitional segments of the individuals in the scene.

Clearly, if the number of o-space centres is unconstrained, the maximum a posteriori probability (MAP) occurs when each individual has his own separate o-space centre, generating a \emph{spurious} F-formation formed by a single individual, that is, $O_{G_i}=TS_i$. To prevent this from happening, we associate a minimum description length prior (MDL) over the number of o-space centres used. This prior takes the same form as dictated by the Akaike Information Criterion (AIC)~\cite{bozdogan1987model}, linearly penalising the log-likelihood for the number of models used.
%
\begin{equation}
\Pr(C=O_{G}|TS)\propto\prod_{i \in [1,n]}\exp\left(-\frac{(u_{G_i}-x_{\mu_i})^2+(v_{G_i}-y_{\mu_i})^2}{\sigma^2} \right) \cdot \exp(-|O_G|)
\end{equation}
where $|O_G|$ is the number of distinct F-formations.

To find the MAP solution, we take the negative log-likelihood and discarding normalising constants, we have the following objective $J(\cdot)$ in standard form:
\begin{equation}
J(O_G|TS)=\sum_{i \in [1,n]}(u_{G_i}-x_{\mu_i})^2+(v_{G_i}-y_{\mu_i})^2 +\sigma^{-2}|O_G|
\label{simplecost}
\end{equation}

As such, this can be seen as optimising a least-squares error combined with an MDL prior. In principle this could be optimised using a standard technique such as k-means clustering combined with a brute force search over all possible choices of $k$ to optimise the MDL cost. In practice, k-means frequently gets stuck in local optima\footnote{In fact, using the technique described the least squares component of the error frequently increases, instead of decreasing, as k increases.} and instead we make use of the graph-cut based optimisation described in~\cite{co-oc}, and widely used in computer vision~\cite{boykov2001graphcuts,lombaert2005graphcuts,campbell2010graphcuts,xu2007graphcuts}

In short, we start from an abundance of possible o-space centres, and then we use a hill-climbing optimisation that alternates between assigning individuals to o-space centres using the efficient graph-cut based optimisation~\cite{co-oc} that directly minimises the cost \eqref{simplecost}, and then minimising the least squares component by updating o-space centres to the mean of $O_g$, for all the individuals $\{i\}$ currently assigned to the F-formation. The whole process is iterated until convergence. This approach is similar to the standard k-means algorithm, sharing both the assignment, and averaging step. However, as the graph-cut algorithm selects the number of clusters, we can avoid local minima by initialising with an excess of model proposals.
In practice, we start from the previously mentioned trivial solution in which each individual is associated with its own o-space centre, centred on his position.

\begin{algorithm}[h!]
  \caption{Finding shared focal centres}
  \begin{algorithmic}
    \STATE Initialise with $O_{G_i} = TS_i \;\;\;\; \forall i \in [1,...,n]$
    \STATE old\_cost$=\infty$
    \WHILE {$J(O_G,TS)<$old\_cost}
    \STATE old\_cost $\leftarrow$ $J(O_G,TS)$
    \STATE run graph cuts to minimise cost \eqref{simplecost}
    \FOR{$\forall g \in [1,...,M]$}
    \IF {$g$ is not empty }
    \STATE update $O_G \leftarrow \frac{\sum_{i\in g} TS_i}{|g|}$
    \ENDIF
    \ENDFOR
    \ENDWHILE
  \end{algorithmic}
\end{algorithm}

\subsection{Visibility constraints}
Finally, we add the natural constraint that people can only join an F-Formation if they can see the o-space centres. By allowing other people to occlude the o-space centre, we are able to capture more subtle nuances such as people being crowded out of F-formations or deliberately ostracised.
Broadly speaking, an individual is excluded from an F-formation when another individual stands between him and the group centre.  Taking $\theta^g_{i,j}$ as the angle between two individuals about a given o-space centre $O_g$ for which is assumed $G_i=G_j=g$ and $d^g_i$, $d^g_j$ as the distance of $i$, or $j$, respectively from the o-space centre $O_g$, the following cost captures this property:
\begin{equation}\label{repulse}
R_{i,j}(g)=\begin{cases}
0 &\text{ if }\theta^g_{i,j}\leq \hat \theta, \text{ or } {d^g_i<d^g_j}\\
\exp\left(K\cos(\theta^g_{i,j})\right)\frac{d^g_i-d^g_j}{d^g_j}&\text{otherwise.}
\end{cases}
\end{equation}
and use the new cost function:
\begin{equation}
\label{full_cost}
J'(O_G|TS)=J(O_G|TS)+\sum_{i,j\in P}R_{i,j}(G_i)
\end{equation}
$R_{i,j}(g_i)$ acts as a visibility constraint on $i$ regardless of the group person $j$ is assigned to, as such it can be treated as a unary cost or data-term and included in the graph-cut based part of the optimisation.
Now we turn to other half of the optimisation - updating the o-space centres. Although, given an assignment of people to a o-space centre, a  local minima can be found using any off the shelf non-convex optimisation, we take a different approach.
There are two points to be aware of: first, the difference between $J'$ and $J$ is sharply peaked and close to zero in most locations, and can generally be safely ignored; second and more importantly, we may often want to move out of a local minima. If updating an o-space centre results in a very high repulsion cost to one individual, this can often be dealt with by assigning the individual to a new group, and this will result in a lower overall cost, and more accurate labelling. As such, when optimising the o-space centres, we pass two proposals for each currently active model to graph-cuts -- the previous proposal generated, and a new proposal based on the current mean of the F-formation. As the graph-cut based optimisation starts from the previous solution, and only moves to lower cost labellings, the cost always decreases and the procedure is guaranteed to converge to a local optimum.

\section{Experiments}
\label{sec:Exp}
The experiments section contains the most exhaustive analysis of the group detection methods in still images carried so far in the computer vision literature, to the best of our knowledge.

In the preliminary part, we describe the five publicly available datasets employed as benchmark, the six methods taken into account as comparison and the metrics adopted to evaluate the detection performances.
Subsequently, we start with an explicative example of how our approach GCFF does work, considering a synthetic scenario taken from the Synthetic dataset.
The experiments continue with a comparative evaluation of GCFF on all the benchmarks against all the comparative methods, looking for the best performance of each approach. Here, GCFF definitely outperforms all the competitors, setting in all the cases new state-of-the-art scores. The ability of detecting groups of a given cardinality and a noise robustness analysis conclude the section, further promoting our technique.


\subsection{Datasets}
\label{sec:datasets}
Five publicly available datasets are used for the experiments: two from \cite{Cristani:FF:BMVC:2011} (\emph{Synthetic} and \emph{Coffee Break}), one from \cite{hung2011detecting} (\emph{IDIAP Poster Data}), one from \cite{Cristani:FF:ICIP:2013} (\emph{Cocktail Party}), and one from \cite{Cristani:SocialExpert:ExpSys:2012} (\emph{GDet}).
A summary of the dataset features is in Table~\ref{tab:datasets}, while a detailed presentation of each dataset follows.  All these datasets are publicly available and the participants to the original experiments gave their permission to share the images and video for scientific purposes. In Fig.~\ref{fig:eximages}, some frames of all the datasets are shown.

\begin{table}[!h]
  \centering
  \footnotesize
  \begin{tabular}{|l|c|c|c|}
    \hline
    \textbf{Dataset}       & \textbf{Data Type} & \textbf{Detection} & \textbf{Detection Quality} \\
    \hline
    \emph{Synthetic}       & synthetic &     --    &      perfect      \\
    \hline
    \emph{IDIAP Poster}    &   real    &  manual   &     very high     \\
    \hline
    \emph{Cocktail Party}  &   real    & automatic &        high       \\
    \hline
    \emph{Coffee Break}    &   real    & automatic &        low        \\
    \hline
    \emph{GDet}            &   real    & automatic &     very low      \\
    \hline
  \end{tabular}
  \vspace{2mm}
  \caption{Summary of the features of the datasets used for experiments.}
  \label{tab:datasets}
\end{table}

\subsubsection*{Synthetic Data\footnote{\url{http://profs.sci.univr.it/{\~}cristanm/datasets.html}}}
\label{sec:synth}
A psychologist generated a set of 10 diverse situations, each one repeated with minor variations for 10 times, resulting in 100 frames representing different social situations, with the aim to span as many configurations as possible for F-formations.
An average of 9 individuals and 3 groups are present in the scene, while there are also individuals not belonging to any group. Proxemic information is noiseless in the sense that there is no clutter in the position and orientation state of each individual.

\subsubsection*{IDIAP Poster Data (IPD)\footnote{\url{http://www.idiap.ch/scientific-research/resources}}}
\label{sec:ipd}
Over 3 hours of aerial videos (resolution $654\times439$px) have been recorded during a poster session of a scientific meeting. Over 50 people are walking through the scene, forming several groups over time. A total of $82$ images were selected with the idea to maximise the crowdedness and variance of the scenes. 
Images are unrelated to each other in the sense that there are no consecutive frames, and the time lag between them prevents to exploit temporal smoothness.
As for the data annotation, a total of 24 annotators were grouped into 3-person subgroups and they were asked to identify F-formations and their associates from static images. Each person's position and body orientation was manually labelled and recorded as pixel values in the image plane -- one pixel represented approximately 1.5cm. The difficulty of this dataset lies in the fact that a great variety of F-formation typologies are present in the scenario (other than circular, L-shapes, side-by-side are present).

\subsubsection*{Cocktail Party (CP)\footnote{The dataset is available by writing to the authors of \cite{lanz:2008:jbt}}}
\label{sec:cp}
This dataset contains about 30 minutes of video recordings of a cocktail party in a $30m^2$ lab environment involving 7 subjects. The party was recorded using four synchronised angled-view cameras (15Hz, $1024\times768$px, jpeg) installed in the corners of the room.
Subject's positions were logged using a particle filter-based body tracker \cite{lanz:2006:pami} while head pose estimation is computed as in \cite{lanz:2008:jbt}.
Groups in one frame every 5 seconds were manually annotated by an expert, resulting in a total of 320 labelled frames for evaluation. This is the first dataset where proxemic information is estimated automatically, so errors may be present. Anyway, due to the highly supervised scenario, errors are very few.

\subsubsection*{Coffee Break (CB)\footnote{\url{http://profs.sci.univr.it/\~cristanm/datasets.html}}}
\label{sec:cb}
The dataset focuses on a coffee-break scenario of a social event, with a maximum of 14 individuals organised in groups of 2 or 3 people each. Images are taken from a single camera with resolution of $1440\times1080$px.
People positions have been estimated by exploiting multi-object tracking on the heads, and head detection has been performed afterwards~\cite{tosato:2013:pami}, considering solely 4 possible orientations (front, back, left and right) in the image plane.
The tracked positions and head orientations were then projected onto the ground plane.
Considering the ground truth data, a psychologist annotated the videos indicating the groups present in the scenes, for a total of 119 frames split in two sequences.
The annotations were generated by analysing each frame in combination with questionnaires that the subjects filled in. This dataset represent one of the most difficult benchmark, since the rough head orientation information, also affected by noise, gives in many cases unreliable information. Anyway, it represents also one of the most realistic scenario, since all the proxemic information comes from automatic, off/the/shelf, computer vision tools.

\subsubsection*{GDet\footnote{\url{http://www.lorisbazzani.info/code-datasets/multi-camera-dataset}}}
\label{sec:gdet}
The dataset is composed by 5 subsequences of images acquired by 2 angled-view low resolution cameras ($352\times328$px) a number of frames spanning from 17 to 132, for a total of 403 annotated frames.
The scenario is a vending machines area where people meet and chat while they are having coffee. This is similar to Coffee Break scenario but in this case the scenario is indoor, which makes occlusions in this case many and severe; moreover, people in this scenario knows each other in advance.
The videos were acquired with two monocular cameras, located on opposite angles of the room. To ensure the natural behaviour of people involved, they were not aware of the experiment purposes.
Ground truth generations follows the same protocol as in Coffee Break; but in this case people tracking has been performed using the particle filter proposed in~\cite{lanz:2006:pami}. Also in this case, head orientation was fixed to 4 angles. This dataset, together with Coffee Break, is the closest to what computer vision can give as input to our a FCG detection technique.

\subsection{Alternative methods}
\label{sec:compmethods}
As alternative methods, we consider all the suitable approaches proposed in the state of the art. 
Six methods are taken into account, one exploiting the concept of \emph{view frustum} (IRPM~\cite{Cristani:SocialExpert:ExpSys:2012}), two approaches based on dominant-sets (DS~\cite{hung2011detecting} and IGD~\cite{tran2013social}) and three different version of Hough Voting approaches using linear accumulator \cite{Cristani:FF:BMVC:2011}, entropic accumulator \cite{setti2013group} and a multi-scale procedure \cite{Cristani:FF:ICIP:2013}. It follows a brief overview of the different methods -- some of them being explained in the Introduction and in the Literature Review section. Please refer to the specific papers for more details about the algorithms.

\subsubsection*{Inter-Relation Pattern Matrix (IRPM)}
\label{sec:irpm}
Proposed by Bazzani et al.~\cite{Cristani:SocialExpert:ExpSys:2012}, it uses the head direction to infer the 3D view frustum as approximation of the Focus of Attention (FoA) of an individual; given the FoA and proximity information, interactions are estimated: the idea is that close-by people whose view frustum is intersecting are in some way interacting. 

\subsubsection*{Dominant Sets (DS)}
\label{sec:ds}
Presented by Hung and Kr\"{o}se~\cite{hung2011detecting}, this algorithm considers an F-formation as a dominant-set cluster~\cite{pavan2007dominant} of an edge-weighted graph, where each node in the graph is a person, and the edges between them measure the affinity between pairs.

\subsubsection*{Interacting Group Discovery (IGD)}
\label{sec:igd}
Presented by Tran et al.~\cite{tran2013social}, it is based on dominant sets extraction from an undirected graph where nodes are individuals and the edges have a weight proportional to how much people are interacting. This method is similar to DS, but it differs in the way the weights of the edges in the graph are computed; in particular, it exploits social cues to compute this weight, approximating the attention of an individual as an ellipse centred at a fixed offset in front of him. Interaction is based on the intersection of the attention ellipses related to two individuals: the more overlap between ellipses, the more they are interacting.

\subsubsection*{Hough Voting for F-formation (HVFF)}
Under this caption, we consider a set of methods based on a Hough Voting strategy to build accumulation spaces and find local maxima of this function.
The general idea is that each individual is associated with a Gaussian probability density function which describes the position of the o-space centre he is pointing at. The pdf is approximated by a set of samples, which basically vote for a given o-space centre location.
The voting space is then quantized and the votes are aggregated on squared cells, so to form a discrete accumulation space. Local maxima in this space identify o-space centres, and consequently, F-formations.
The first work in this field is~\cite{Cristani:FF:BMVC:2011}, where the votes are linearly accumulated by just summing up all the weights of votes belonging to the same cell. A first improvement of this approach is presented in~\cite{setti2013group}, where the votes are aggregated by using the weighted Boltzmann entropy function. In~\cite{Cristani:FF:ICIP:2013} a multi-scale approach is used on top of the entropic version: the idea is that groups with higher cardinality tends to arrange around a larger o-space; the entropic group search runs for different o-space dimensions by filtering groups cardinalities; afterwards, a fusion step is based on a majority criterion.

\subsection{Evaluation metrics}
\label{sec:metrics}
As accuracy measures, we adopt the metrics proposed in \cite{Cristani:FF:BMVC:2011} and extended in \cite{setti2013group}: we consider a group as correctly estimated if at least $\lceil(T \cdot |G|)\rceil$ of their members are found by the grouping method and correctly detected by the tracker, and if no more than $1-\lceil(T \cdot |G|)\rceil$ false subjects (of the detected tracks) are identified, where $|G|$ is the cardinality of the labelled group $G$, and $T \in\; ]0,1]$ is an arbitrary threshold, called \emph{tolerance threshold}.
In particular, we focus on two interesting values of $T$: $2/3$ and $1$.

With this definition of \emph{tolerant match}, we can determine for each frame the correctly detected groups (true positives -- TP), the miss-detected groups (false negatives -- FN) and the hallucinated groups (false positives -- FP). With this, we compute the standard pattern recognition metrics precision and recall:
\begin{equation}
  precision = \frac{TP}{TP + FP}  \;\; , \;\;\;\;  recall = \frac{TP}{TP + FN}
\end{equation}
and the $F_1$ score defined as the harmonic mean of precision and recall:
\begin{equation}
  F_1=2\cdot\frac{precision\,\cdot\, recall}{precision\, +\, recall}
\end{equation}

In addition to these metrics, we present in this paper a new metric which is independent from the tolerance threshold $T$. We compute this new score as the area under the curve (AUC) in the $F_1$ \textsc{vs.} $T$ graph with $T$ varying from 1/2 to 1\footnote{Please note that we avoid to consider $0<T<1/2$, since in this range we are accepting as good those groups where more than the half of the subjects is missing or false positive, resulting in useless estimates.}.
We will call it \emph{Global Tolerant Matching} score (GTM).
Since in our experiments we only have groups up to 6 individuals, without loss of generality we consider $T$ varying with 3 equal steps in the range stated above.

Moreover, we will discuss results also in terms of group cardinality, by computing the $F_1$ score for each cardinality separately and then computing mean and standard deviation.

\subsection{An explicative example}
\begin{figure}[!h]
  \centering
  \includegraphics[width=.9\columnwidth]{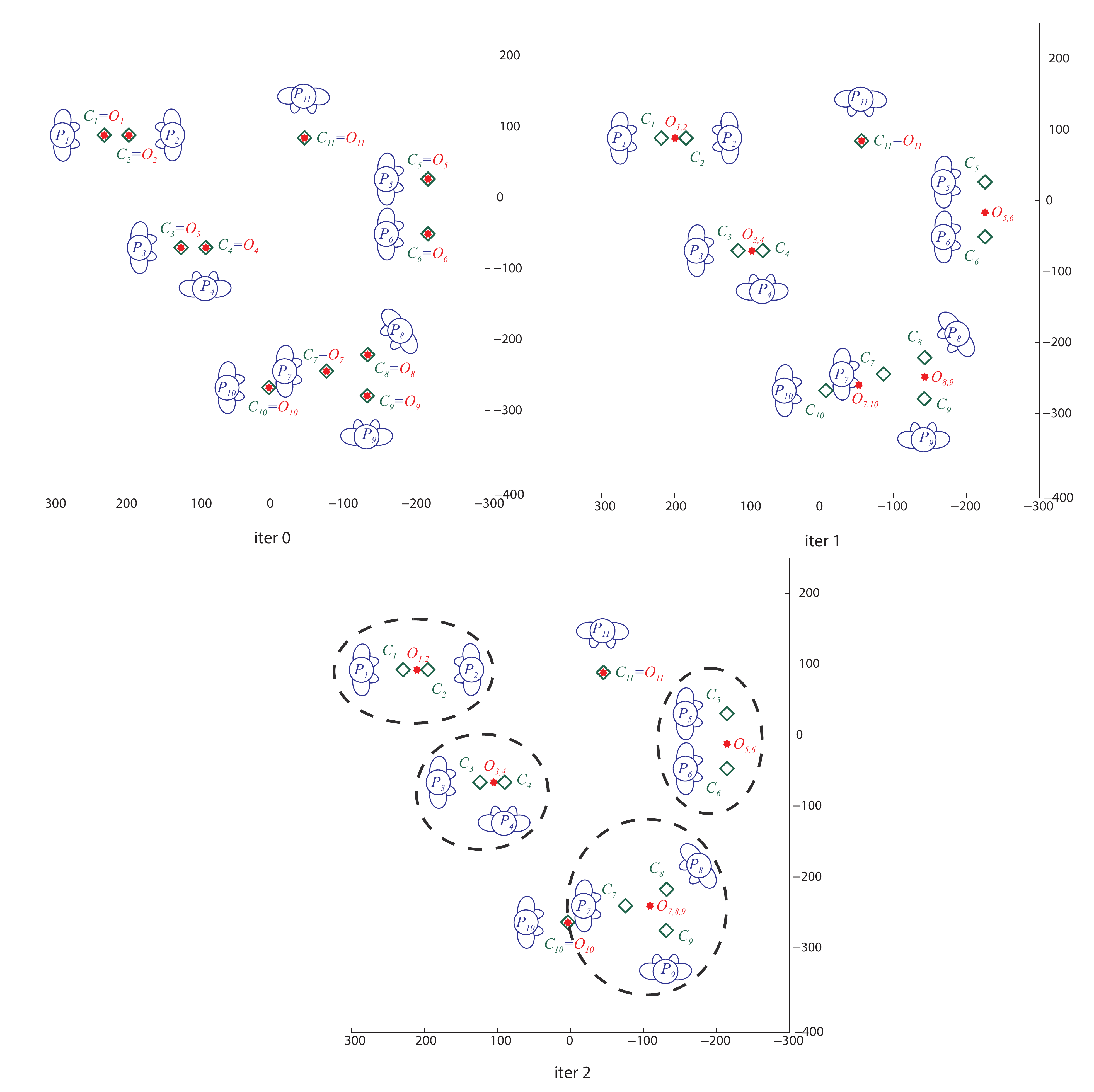}
  \caption{{\bf An explicative example.} Iteration 0: initialization with the candidate o-space centres $\{O\}$ coincident with the transitional segment of each individual $\{C\}$. Iteration 1: first graph-cuts run; easy groups are correctly clustered while the most complex still present errors (the FCG formed by $P_7$ and $P_20$ violates the visibility constraint). Iteration 2: the second graph-cuts run correctly detects the $O_{7,8,9}$ F-formation (at the bottom). Se text for more details.}
  \label{fig:exrun}
\end{figure}

Figure~\ref{fig:exrun} gives a visual insight of our graph-cuts process.
Given the position and orientation of each individual $P_i$, the algorithm starts by computing the transitional segments $C_i$.
At the first iteration $0$, the candidate o-space centres $O_i$ are initialized, and are coincident with the transitional segments $C_i$; in this example are present 11 individuals, so 11 candidate o-space centres are generated.
After iteration $1$, the proposed segmentation process provides 1 singleton ($P_{11}$) and 5 FCGs of two individuals each. We can appreciate different configurations such as  \emph{vis-a-vis} ($O_{1,2}$), L-shape ($O_{3,4}$) and side-by-side ($O_{5,6}$). Still, the grouping in the bottom part of the image is wrong ($P_7$ to $P_{10}$), since it violates the exclusion principle.
In iteration 2, the previous candidate o-space centres is considered as initialization, and a new graph is built. In this new configuration, the group $O_{7,10}$ is recognized as violating the visibility constraint and thus the related edge is penalized; a new run of graph-cuts minimization allows to correctly cluster the FCGs in a singleton ($P_{10}$) and a FCG formed by three individuals ($O_{7,8,9}$), which corresponds to the ground truth (visualized as the dashed circles).

\subsection{Best results analysis}
\label{sec:best}
Given the metrics explained above, the first test analyses the best performances for each method on each dataset; in practice, a tuning phase has been carried out for each method$/$dataset combination in order to get the best performances\footnote{We did not have code for Dominant Sets~\cite{hung2011detecting} and thus we used results provided directly from the authors of the method for a subset of data. For this reason, average results over all the datasets are only averaged over 3 datasets, and cannot be taken into account for a fair comparison.\label{foot:ds}}. Best parameters (found on half of one sequence by cross-validation, and kept unchanged for the remaining datasets) are reported in Table~\ref{tab:params}. Please note, finding the right parameters can also fixed by hand, since the stride $D$ depends on the social context under analysis (formal meetings will have higher $D$, the presence of tables and similar items may also increase the diameter of the FCGs): with a given $D$, for example, it is assumed that circular F-formations will have diameter of $2D$. The parameter $\sigma$ indicates how much we are permissive in accepting deviations from such a diameter. Moreover, $D$ depends also on the different measure units (pixels/cm) which characterize the proxemic information associated to each individual in the scene.

\begin{table}[!h]
  \centering
  \begin{tabular}{|l|c|c|c|}
    \hline
    \textbf{Dataset}       & stride $D$ & \emph{std} $\sigma$  \\
    \hline
    \emph{Synthetic}       & 30 & 80 \\ 
    \hline
    \emph{IDIAP Poster}    & 20 & 45 \\ 
    \hline
    \emph{Cocktail Party}  & 70 & 170 \\
    \hline
    \emph{Coffee Break}    & 30 & 85 \\
    \hline
    \emph{GDet}            & 30 & 200 \\ 
    \hline
  \end{tabular}
  \vspace{2mm}
  \caption{Parameters used in the experiments for each dataset. These parameters are the results of a tuning phase and the difference are due to different measure units (pixels/cm) and different social environments (indoor/outdoor, formal/informal, etc.).}
  \label{tab:params}
\end{table}

Table~\ref{tab:best_t06} shows best results by considering the threshold $T=2/3$, which corresponds to find at least $2/3$ of the members of a group, no more than $1/3$ of false subjects; while Table~\ref{tab:best_t1} presents results with $T=1$, considering a group as correct if all and only its members are detected. The proposed method outperforms all the competitors, on all the datasets. With $T=2/3$, three observations can be made: the first is that our approach GCFF improves substantially the precision (of $13\%$ in average) and even more definitely  the recall scores  (of $17\%$ in average) of the state of the art approaches. The second is that our approach produces the same score for both the precision and the recall; this is very convenient and convincing, since so far all the approaches of FCG detections have shown to be weak in the recall dimension. The third observation is that GCFF performs well both in the case where no errors in the position or orientation of the people are present (as the Synthetic dataset) and in the cases where strong noise of position and orientation  is present (Coffee Break, GDet).

When moving to tolerance threshold equal to $1$ (all the people in a group have to be individuated, and no false positive are allowed) the performance is reasonably lower, but the increment is even stronger w.r.t. to the state of the art, in general on all the datasets: in particular, on the Cocktail Party dataset, the results are more than twice the scores of the competitors. Finally, even in this case,  GCFF produces a very similar score for precision and recall.

\begin{table*}
  \centering
  \scalebox{0.7}{\hspace{-3.5cm}
  \begin{tabular}{|l|c|c|c|c|c|c|c|c|c|c|c|c|c|c|c|c|c|c|}
    \hline
     & \multicolumn{3}{|c|}{\emph{Synthetic}} & \multicolumn{3}{|c|}{\emph{IDIAP Poster}} &
       \multicolumn{3}{|c|}{\emph{Cocktail Party}} & \multicolumn{3}{|c|}{\emph{Coffee Break}} &
       \multicolumn{3}{|c|}{\emph{GDet}} & \multicolumn{3}{|c|}{\emph{\textbf{Total}}} \\
    \hline
     & prec. & rec. & $F_1$ & prec. & rec. & $F_1$ & prec. & rec. & $F_1$ & prec. & rec. & $F_1$ & prec. & rec. & $F_1$ & prec. & rec. & $F_1$ \\
    \hline
    IRPM~\cite{Cristani:SocialExpert:ExpSys:2012}    & 0.85 & 0.80 & 0.82 & 0.82 & 0.74 & 0.78 & 0.56 & 0.43 & 0.49 & 0.68 & 0.50 & 0.57 & 0.77 & 0.47 & 0.58 & 0.70 & 0.49 & 0.56 \\
    DS~\cite{hung2011detecting}           & 0.85 & 0.97 & 0.90 & 0.91 & 0.92 & 0.91 &  --  &  --  &  --  & 0.69 & 0.65 & 0.67 &  --  &  --  &  --  & \it{0.81}\footref{foot:ds} & \it{0.83}\footref{foot:ds} & \it{0.82}\footref{foot:ds} \\
    IGD~\cite{tran2013social}        & 0.95 & 0.71 & 0.81 & 0.80 & 0.68 & 0.73 & 0.81 & 0.61 & 0.70 & 0.81 & 0.78 & 0.79 & 0.83 & 0.36 & 0.50 & 0.68 & 0.76 & 0.70 \\
    \hline
    HVFF lin~\cite{Cristani:FF:BMVC:2011} & 0.75 & 0.86 & 0.80 & 0.90 & 0.95 & 0.92 & 0.59 & 0.74 & 0.65 & 0.73 & 0.86 & 0.79 & 0.66 & 0.68 & 0.67 & 0.75 & 0.79 & 0.76 \\
    HVFF ent~\cite{setti2013group}  & 0.79 & 0.86 & 0.82 & 0.86 & 0.89 & 0.87 & 0.78 & 0.83 & 0.80 & 0.76 & 0.86 & 0.81 & 0.69 & 0.71 & 0.70 & 0.78 & 0.78 & 0.77 \\
    HVFF ms~\cite{Cristani:FF:ICIP:2013}     & 0.90 & 0.94 & 0.92 & 0.87 & 0.91 & 0.89 & 0.81 & 0.81 & 0.81 & 0.83 & 0.76 & 0.79 & 0.71 & 0.73 & 0.72 & 0.84 & 0.66 & 0.74 \\
    \hline
    \bf GCFF                            & \bf0.97 & \bf0.98 & \bf0.97 & \bf0.94 & \bf0.96 & \bf0.95 & \bf0.84 & \bf0.86 & \bf0.85 & \bf0.85 & \bf0.91 & \bf0.88 & \bf0.92 & \bf0.88 & \bf0.90 & \bf0.89 & \bf0.89 & \bf0.89 \\
    \hline
  \end{tabular}
  }
  \caption{Average precision, recall and $F_1$ scores for all the methods and all the datasets. ($T=2/3$)}
  \label{tab:best_t06}
  \vspace{1em}
  %
  \scalebox{0.7}{\hspace{-3.5cm}
  \begin{tabular}{|l|c|c|c|c|c|c|c|c|c|c|c|c|c|c|c|c|c|c|}
    \hline
     & \multicolumn{3}{|c|}{\emph{Synthetic}} & \multicolumn{3}{|c|}{\emph{IDIAP Poster}} &
       \multicolumn{3}{|c|}{\emph{Cocktail Party}} & \multicolumn{3}{|c|}{\emph{Coffee Break}} &
       \multicolumn{3}{|c|}{\emph{GDet}} & \multicolumn{3}{|c|}{\emph{\textbf{Total}}} \\
    \hline
     & prec. & rec. & $F_1$ & prec. & rec. & $F_1$ & prec. & rec. & $F_1$ & prec. & rec. & $F_1$ & prec. & rec. & $F_1$ & prec. & rec. & $F_1$ \\
    \hline
    IRPM~\cite{Cristani:SocialExpert:ExpSys:2012}    & 0.53 & 0.47 & 0.50 & 0.71 & 0.64 & 0.67 & 0.28 & 0.17 & 0.21 & 0.27 & 0.23 & 0.25 & 0.59 & 0.29 & 0.39 & 0.46 & 0.29 & 0.35 \\
    DS~\cite{hung2011detecting}        & 0.68 & 0.80 & 0.74 & 0.79 & 0.82 & 0.81 &  --  &  --  &  --  & 0.40 & 0.38 & 0.39 &  --  &  --  &  --  & \it{0.60}\footref{foot:ds} & \it{0.63}\footref{foot:ds} & \it{0.62}\footref{foot:ds} \\
    IGD~\cite{tran2013social}        & 0.30 & 0.22 & 0.25 & 0.31 & 0.27 & 0.29 & 0.23 & 0.10 & 0.13 & 0.50 & 0.50 & 0.50 & 0.67 & 0.20 & 0.31 & 0.45 & 0.21 & 0.27 \\
    \hline
    HVFF lin~\cite{Cristani:FF:BMVC:2011} & 0.64 & 0.73 & 0.68 & 0.80 & 0.86 & 0.83 & 0.26 & 0.27 & 0.27 & 0.41 & 0.47 & 0.44 & 0.43 & 0.45 & 0.44 & 0.43 & 0.46 & 0.44 \\
    HVFF ent~\cite{setti2013group}  & 0.47 & 0.52 & 0.49 & 0.72 & 0.74 & 0.73 & 0.28 & 0.30 & 0.29 & 0.47 & 0.52 & 0.49 & 0.44 & 0.45 & 0.45 & 0.42 & 0.44 & 0.43 \\
    HVFF ms~\cite{Cristani:FF:ICIP:2013}     & 0.72 & 0.73 & 0.73 & 0.73 & 0.76 & 0.74 & 0.30 & 0.30 & 0.30 & 0.40 & 0.38 & 0.39 & 0.44 & 0.45 & 0.45 & 0.44 & 0.45 & 0.45 \\
    \hline
    \bf GCFF                           & \bf0.91 & \bf0.91 & \bf0.91 & \bf0.85 & \bf0.87 & \bf0.86 & \bf0.63 & \bf0.65 & \bf0.64 & \bf0.61 & \bf0.64 & \bf0.63 & \bf0.73 & \bf0.68 & \bf0.71 & \bf0.71 & \bf0.70 & \bf0.71 \\
    \hline
  \end{tabular}
  }
  \caption{Average precision, recall and $F_1$ scores for all the methods and all the datasets. ($T=1$)}
  \label{tab:best_t1}
\end{table*}

\begin{figure}[!h] 
  \centering
  \includegraphics[width=.8\columnwidth]{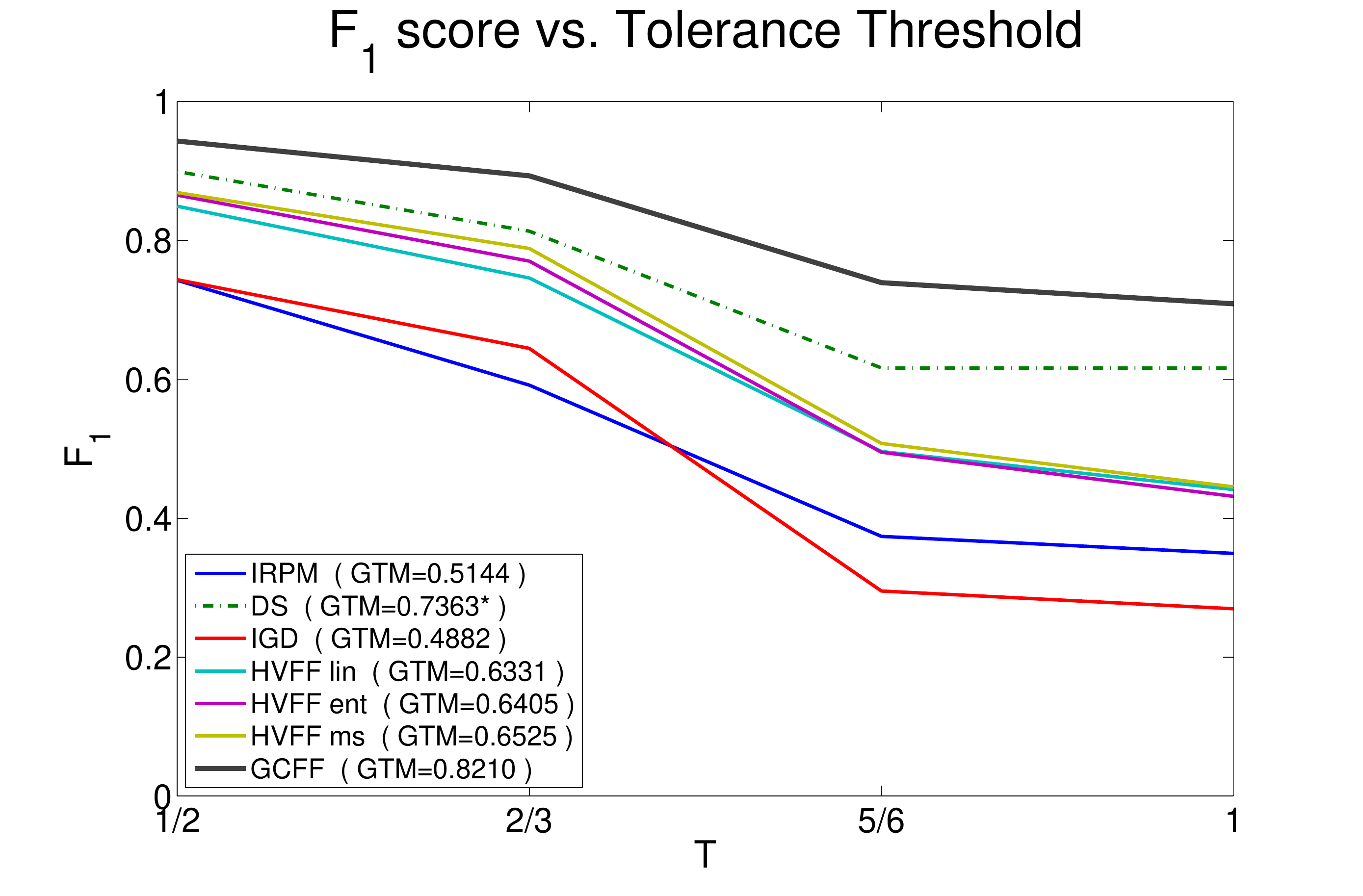}
  \caption{{\bf Global $F_1$ score \textsc{vs.} tolerance threshold $T$.} Between brackets in legend the Global Tolerant Matching score. Dominant Sets (DS) is averaged over 3 datasets only, because of results availability. (Best viewed in colour)}
  \label{fig:thvar}
\end{figure}

A performance analysis is also provided by changing the tolerance threshold $T$. Fig.~\ref{fig:thvar} shows the average $F_1$ scores for each method computed over all the frames and datasets. From the curves we can appreciate how the proposed method is consistently best performing for each $T$-value. In the legend of Fig.~\ref{fig:thvar} the Global Tolerant Matching score is also reported. Again, GCFF is outperforming the state of the art, independently from the choice of $T$.

The reason why our approach does better than the competitors  has been explained in the state of the art section, here briefly summarized: the Dominant Set-based approaches DS and IGD, even if they are based on an elegant optimization procedure, tend to find circular groups, and are weaker in individuating other kinds of F-formations. Hough-based approaches HVFF X (X= \emph{lin}, \emph{ent}, \emph{ms}) have a good modeling of the F-formation, allowing to find any shape, but rely on a greedy optimization procedure. Finally, IRPM approach has a rough modeling of the F-formation. Our approach viceversa has a rich modeling of the F-formation, and a powerful optimization strategy.

\subsection{Cardinality analysis}
\label{sec:cardinality}
As stated in~\cite{Cristani:FF:ICIP:2013}, some methods are shown to work better with some group cardinalities. In this experiment, we sistematically check this aspect, evaluating the performance of all the considered methods in individuating groups with a particular number of individuals.
Since Synthetic, Coffee Break and IDIAP Poster Session datasets only have groups of cardinality 2 and 3, we only focus on the remaining 2 datasets, which have a more uniform distribution of groups cardinalities.
Tables~\ref{tab:card_CP} and \ref{tab:card_GDet} show $F_1$ scores for each method and each group cardinality respectively for Cocktail Party and GDet datasets.
In both cases the proposed method outperforms the other state of the art methods in terms of higher average $F_1$ score, with very low standard deviation. In particular, only IRPM gives in GDet dataset results which are more stable than ours, but they are definitely poorer.

\begin{table*}
  \centering
  \footnotesize
  \begin{tabular}{|l|c|c|c|c|c|c|c|}
    \hline
     & $k=2$ & $k=3$ & $k=4$ & $k=5$ & $k=6$ & \bf{Avg} & \bf{Std} \\
    \hline
    $\#$ groups & 81 & 82 & 44 & 55 & 147 & -- & -- \\
    \hline\hline
    IRPM~\cite{Cristani:SocialExpert:ExpSys:2012}    & 0.26 & 0.53 & 0.74 & 0.42 & 0.59 & 0.51 & 0.18\\
    IGD~\cite{tran2013social}        & 0.06 & 0.52 & 0.66 & 0.73 & 0.85 & 0.56 & 0.30\\
    \hline
    HVFF lin~\cite{Cristani:FF:BMVC:2011} & 0.38 & \it{0.76} & 0.57 & 0.67 & 0.94 & 0.66 & 0.21\\
    HVFF ent~\cite{setti2013group}  & 0.45 & 0.75 & 0.69 & 0.73 & \it{0.96} & 0.71 & 0.18\\
    HVFF ms~\cite{Cristani:FF:ICIP:2013}     & 0.49 & 0.74 & 0.70 & 0.71 & \it{0.96} & 0.72 & 0.17\\
    \hline\hline
    \bf GCFF                     & \it{0.59} & 0.64 & \it{0.80} & \it{0.85} & 0.94 & \bf{0.76} & \bf{0.14} \\
    \hline
  \end{tabular}
  \caption{Cocktail Party -- $F_1$ score \textsc{vs.} cardinality. ($T=1$)}
  \label{tab:card_CP}
  \vspace{2em}
  \begin{tabular}{|l|c|c|c|c|c|c|c|}
    \hline
     & $k=2$ & $k=3$ & $k=4$ & $k=5$ & $k=6$ & \bf{Avg} & \bf{Std} \\
    \hline
    $\#$ groups & 197 & 124 & 22 & 35 & 13 & -- & -- \\
    \hline\hline
    IRPM~\cite{Cristani:SocialExpert:ExpSys:2012}    & 0.40 & 0.59 & 0.45 & 0.42 & 0.35 & 0.44 & \bf{0.09}\\
    IGD~\cite{tran2013social}        & 0.15 & 0.52 & 0.33 & 0.54 & 0.83 & 0.47 & 0.25\\
    \hline
    HVFF lin~\cite{Cristani:FF:BMVC:2011} & 0.51 & 0.76 & 0.03 & 0.16 & 0.13 & 0.32 & 0.31\\
    HVFF ent~\cite{setti2013group}  & 0.57 & 0.73 & 0.24 & 0.23 & 0.13 & 0.38 & 0.26\\
    HVFF ms~\cite{Cristani:FF:ICIP:2013}     & 0.56 & 0.78 & 0.17 & 0.41 & 0.67 & 0.52 & 0.23\\
    \hline\hline
    \bf GCFF                     & \it{0.74} & \it{0.87} & \it{0.53} & \it{0.77} & \it{0.88} & \bf{0.76} & 0.14 \\
    \hline
  \end{tabular}
  \caption{GDet-- $F_1$ score \textsc{vs.} cardinality. ($T=1$)}
  \label{tab:card_GDet}
\end{table*}

\subsection{Noise analysis}
\label{sec:noise}
In this experiment, we show how the methods behave against different degrees of clutter.
For this sake, we consider the Synthetic dataset as starting point and we add to the proxemic state of each individual of each frame some random values based on a known noise distribution.
We assume that the noise follows a Gaussian distribution with mean $0$, and noise on each dimension (position, orientation) is uncorrelated. For our experiments we used $\sigma_x = \sigma_y = 20$cm and $\sigma_{\theta} = 0.1$rad.
In our experiments, we consider 11 levels of noise $L_n = 0,\ldots,10$, where
\begin{equation}
  \begin{cases}
        x_n(L_n) = x + \text{randsample} ( \mathcal{N}(0,L_n*\sigma_x) ) \\
        y_n(L_n) = y + \text{randsample} ( \mathcal{N}(0,L_n*\sigma_y) ) \\
        \theta_n(L_n) = \theta + \text{randsample} ( \mathcal{N}(0,L_n*\sigma_{\theta}) )
  \end{cases}
\end{equation}

\begin{figure*}[!h]
  \centering
  \includegraphics[width=\textwidth]{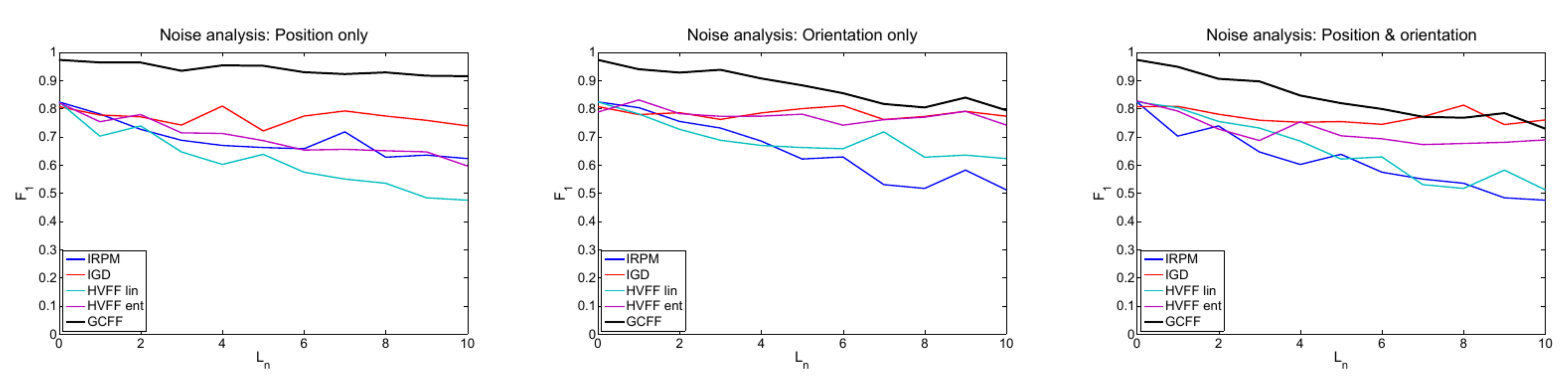}
  \caption{{\bf Noise analysis.} $F_1$ score \textsc{vs.} Noise Level on position (left), orientation (centre) and combined (right). (Best viewed in colour)}
  \label{fig:noise}
\end{figure*}

In particular, we produce results by adding noise on position only (leaving the orientation at its exact value), on orientation only (leaving the position of each individual at its exact value) and on both position and orientation.
Fig.~\ref{fig:noise} shows $F_1$ scores for each method while increasing the noise level.
In this case we can appreciate that with high orientation and combined noise IGD performs comparably or better than GCFF; this is a confirmation of the fact that methods based on Dominant Sets are performing very well when the orientation information is not reliable, as already stated in~\cite{setti2013group}.

\section{Conclusions} \label{sec:Conc}
In this paper we presented a statistical framework for the detection of free-standing conversational groups (FCG) in still images. FCGs represent very common and crucial social events, where social ties (intimate \textsc{vs.} formal relationships) pop out naturally; for this reason,  detection of FCGs is of primary importance in a wide spectra of application. The proposed algorithm is based on a graph-cuts minimization scheme, which essentially clusters individuals into groups; in particular, the computational model implements the sociological definition of F-formation, describing how people forming a FCG will locate in the space. The take-home message is that having basic proxemic information (people location and orientation) is enough to individuate groups with high accuracy. This claim originates from one of the most exhaustive experimental session implemented so far on this matter, with 5 diverse datasets taken into account, and all the best approaches in the literature considered as competitors; in addition to this, a deep analysis on the robustness to noise and on the capability of individuating groups of a given cardinality have been also carried out.
The natural extension of this study consists in analyzing the temporal information, that is, video sequences: in this scenario, interesting phenomena such as entering or exiting a group could be considered and modeled, and the temporal smoothness can be exploited to generate even more precise FCG detections.
%

\section*{Acknowledgments}
F.~Setti and C.~Bassetti are supported by the \textsc{VisCoSo} project grant, financed by the Autonomous Province of Trento through the ``Team 2011'' funding programme.
Portions of the research in this paper used the Idiap Poster Data Corpus made available by the Idiap Research Institute, Martigny, Switzerland.

\bibliographystyle{plain}
\bibliography{mybib}

\end{document}